\documentclass[]{spie}  

\usepackage{graphicx}  
\usepackage{float}     
\usepackage{booktabs}  
\usepackage{array}
\usepackage{amsmath,amsfonts,amssymb}
\usepackage{pifont}
\usepackage{graphicx}
\usepackage[pagebackref=true,breaklinks=true,colorlinks,bookmarks=false]{hyperref}
\usepackage[nameinlink]{cleveref}

\usepackage[T1]{fontenc}
\usepackage{adjustbox}
\usepackage{multirow}
\usepackage{mathtools}
\usepackage{graphicx}
\usepackage[dvipsnames]{xcolor}
\usepackage[title]{appendix}%
\title{Multi-illuminant Color Constancy via Multi-scale Illuminant Estimation and Fusion}

\author[a]{Hang Luo}
\author[a]{Rongwei Li}
\author[a]{Jinxing Liang}

\affil[a]{School of Computer Science and Artificial Intelligence, Wuhan Textile University, No.1 Sunshine Avenue, Jiangxia District, Wuhan, Hubei Province, China}

\authorinfo{Corresponding author: Jinxing Liang (\texttt{jxliang@wtu.edu.cn})}

\begin{document} 
\maketitle

\begin{abstract}
Multi-illuminant color constancy methods aim to eliminate local color casts within an image through pixel-wise illuminant estimation. Existing methods mainly employ deep learning to establish a direct mapping between an image and its illumination map, which neglects the impact of image scales. To alleviate this problem, we represent an illuminant map as the linear combination of components estimated from multi-scale images. Furthermore, we propose a tri-branch convolution networks to estimate multi-grained illuminant distribution maps from multi-scale images. These multi-grained illuminant maps are merged adaptively with an attentional illuminant fusion module. Through comprehensive experimental analysis and evaluation, the results demonstrate the effectiveness of our method, and it has achieved state-of-the-art performance.
\end{abstract}



\section{Introduction}
\label{sec1:intro}
The visual system of human beings can maintain relatively stable color perception even under varying illuminations, and this ability is called color constancy \cite{foster_color_2011}. However, cameras inherently don't possess this ability, which results in reddish or blueish color casts in an image. Color constancy method aims to mimic the visual color constancy of human beings to discount the color cast within an image. It is typically used to fulfill the built-in auto white balance step in a camera \cite{Ramanath2005}, which can significantly enhance image quality and improve the illuminant robustness of downstream vision tasks \cite{afifi_what_2019}.

The mechanism underlying visual color constancy is not yet fully understood. To simplify the problem, existing works approximately simulate the visual color constancy by transforming an image from arbitrary illuminants to canonical white light using von Kries chromatic adaptation transform \cite{hunt_chromatic_2005}. Specifically, it involves two critical steps: (1) estimating the chromaticity of the illuminant, i.e. the illuminant color, and (2) correcting image color with diagonal transform. The key problem is how to accurately estimate the illuminant color, and existing works are dedicated to solve this problem.

A large number of studies further simplify the problem by assuming that there is only one illuminant in the scene, so they are called single-illuminant color constancy method. Classical single-illuminant methods impose additional assumptions on the scene and use image statistics for illuminant estimation \cite{Buchsbaum1980a, barnard_comparison_2002,finlayson_shades_2004,VandeWeijer2007c,gijsenij_improving_2012,Yang2015,qian_finding_2019}. Another line of research has incorporated machine learning techniques into the field, such as neural networks \cite{cardei_estimating_2002}, k-nearest neighbor \cite{gijsenij_color_2007} and Bayesian estimation \cite{Gehler2008}. In recent years, deep learning has revolutionized the field, enabling diverse methods based on convolutional neural networks \cite{Bianco2015,Barron2017,Hu2017, woo_deep_2021,afifi_deep_2020,luo_estimating_2021}, contrastive learning \cite{lo_clcc_2021}, transfer learning \cite{tang_transfer_2022}, etc. Despite these advancements, single-illuminant methods face an inevitable limitation that their assumption is rarely satisfied in practice. Natural scenes typically contain multiple illuminants and single-illuminant methods cannot simultaneously correct all local color casts brought by distinct illuminants.

Multi-illuminant color constancy methods relax the single-illuminant assumption and aim to estimate the illuminant pixel-wisely, resulting in an illuminant map. An intuitive idea is segmenting an image into regions with distinct illuminants and estimating the illuminant for each region \cite{gijsenij_color_2012, beigpour_multi-illuminant_2014, joze_exemplar-based_2014, bianco_single_2017, gao_combining_2019}, but these methods are limited to coarse-grained local illuminant estimation. In recent years, owing to the emergence of large scale multi-illuminant datasets \cite{kim_large_2021}, many methods have used deep learning to achieve fine-grained pixel-wise illuminant estimation \cite{domislovic_color_2023, li_mimt_2023, yue_robust_2024, entok_pixel-wise_2024, kim_attentive_2024}. We can suppose that the illuminant distribution is more uniform at small scales and more diverse in large scales. However, current methods have overlooked the scale-dependent variation of illuminant distribution, consequently limiting their ability to effectively capture diverse illuminant distribution features for precise pixel-wise illuminant estimation.

In this paper, we investigate the impact of image scales and propose a coarse-fine-decomposed framework for pixel-wise illuminant estimation. Specifically, we regard an illuminant map as the linear combination of multi-grained components and build a convolutional framework with three branches, each using an U-Net \cite{navab_u-net_2015} to estimate the illuminant map from an image of a certain scale. Since larger images provide finer details, these branches can capture multi-grained illuminant distribution features, resulting in multi-grained illuminant maps. At the final stage, an attentional illuminant fusion module adaptively integrates these multi-grained illuminant maps by generating pixel-wise weight maps and performing a linear combination. Our contributions are summarized as follows:

(1) We propose that an illuminant map can be decomposed into a group of multi-grained components and propose a framework to estimate multi-grained illuminant maps from multi-scale images;

(2) To adaptively identify and enhance the most relevant illuminant for each pixel, we construct an attentional illuminant fusion module to automatically assign pixel-wise weights for these illuminant maps estimated from multi-scale images;

(3) We conduct extensive experiments to validate and analyze our method, and the experimental results demonstrate the effectiveness of our method.

\section{Related works}
\label{sec2:related works}

\subsection{Single-illuminant Color Constancy}
\label{sec2.1:Single-illuminant Color Constancy}
Existing single-illuminant color constancy methods can be roughly categorized into statistics-based and learning-based methods. Statistics-based methods use the statistics of an image as the illuminant color. Gray World (GW) \cite{Buchsbaum1980a}, one of the most well-known statistics-based methods, assumes that the average reflectance of a scene is gray and employs the mean value of an image as estimation. White Patch (WP) \cite{barnard_comparison_2002} supposes that there is a white patch in the scene and uses the max value of each channel as the illuminant color. Shade of Gray (SoG) \cite{finlayson_shades_2004} uses the p-norm of an image as estimation. Gray Edge (GE) \cite{VandeWeijer2007c} extends SoG
to the gradient domain. Weighted GE \cite{gijsenij_improving_2012} further takes into account the diversity of image edges and proposes an adaptive weighting scheme. Cheng et al. \cite{Cheng2014a} propose to use the first principal component of bright and dark pixels as the estimation. Since gray pixels can directly indicate the illuminant color, Yang et al. \cite{Yang2015} and Qian et al. \cite{qian_finding_2019} respectively propose a method for gray pixel detection. Statistics-based methods are efficient and interpretable, but their generalization ability is limited due to their inherent assumptions.

Learning-based methods estimate the illuminant color in an data-driven manner. Cardei et al. \cite{cardei_estimating_2002} build the mapping between the binarized chromatic histogram and the illuminant color using neural networks. Gijsenij et al. \cite{gijsenij_color_2007} employs k-nearest neighbor clustering in the Weibull parameter space to determine the most appropriate statistics-based method for an image. Gehler et al. \cite{Gehler2008} use Bayesian method for illuminant estimation. Nowadays, deep learning for illuminant estimation is the main direction. Bianon et al. \cite{Bianco2015} build the first convolutional neural networks for illuminant estimation. Barron et al. \cite{Barron2017} reduce the regression problem to a problem of locating the illuminant in the log-chroma space. Hu et al. \cite{Hu2017} build a fully convolutional network. Woo et al. \cite{woo_deep_2021} integrate the dichromatic reflection model with convolutional neural networks, establishing a framework to predict dichromatic planes for illuminant estimation. Afifi et al. \cite{afifi_when_2019, afifi_deep_2020} and Luo et al. \cite{luo_estimating_2021} propose to correct color biases in sRGB color space. In addition to the end-to-end paradigm, other deep learning paradigms have also been explored. Lo et al. \cite{lo_clcc_2021} create illuminant-related contrastive pairs for contrastive learning. Tang et al. \cite{tang_transfer_2022} bridge the gaps between different image domains by transfer learning, expanding the number of available data.

\subsection{Multi-illuminant Color Constancy}
\label{sec2.2:Multi-illuminant Color Constancy}
As mentioned before, multi-illuminant color constancy methods aim to estimate the illuminant color per pixel, and current methods can be classified into two partially overlapping categories: segmentation-based methods and deep-learning-based methods. Gijsenij et al. \cite{gijsenij_color_2012} propose to segment an image into small patches and perform single-illuminant estimation and clustering for illuminant distribution estimation. Beigpour et al. \cite{beigpour_multi-illuminant_2014} use conditional random fields to optimize the initial patch-wise local estimation. Hamid Reza Vaezi \cite{joze_exemplar-based_2014} segments an image into different surfaces according to the rule of color consistency and estimates the illuminant for each surface by finding its nearest neighbor surfaces from the training data. Bianco et al. \cite{bianco_single_2017} use convolutional neural networks for patch-wise illuminant estimation. Gao et al. \cite{gao_combining_2019} segment an image into bright and dark areas, and generate illuminant estimation per area with Difference of Gaussian operators.

For a long time, the development of deep-learning-based methods faces significant challenges due to the absence of large-scale datasets. Therefore, Kim et al. \cite{kim_large_2021} create a large scale multi-illuminant (LSMI) dataset and train an U-Net for illuminant distribution estimation (LSMI-U). Domislović et al. \cite{domislovic_color_2023} build a lightweight convolutional network for patch-wise estimation. Li et al. \cite{li_mimt_2023} propose a multi-task training framework to integrate related tasks into illuminant distribution estimation. Entok et al. \cite{entok_pixel-wise_2024} incorporate smoothing constraints into the training of deep models. Kim et al. \cite{kim_attentive_2024} employ slot attention to decouple the original task into illuminant chromaticity estimation and weight map estimation. Yue et al. \cite{yue_robust_2024} have revealed that some methods are sensitive to image bit-depth and propose a physical-constrained method. Our method differs mainly in that we have taken into account the impact of image scales and use dedicated models to restore multi-grained illuminant distribution maps from multi-scale images.

\section{Method}
\label{sec3: method}
\subsection{Overview}
\label{sec3.1: method overview}
Based on the fact that larger images contain more details, we believe large scale is suitable for fine-grained estimation and small scale for coarse-grained estimation. Thus, we propose to estimate multi-grained illuminant distributions from multi-scale images and use their linear combination as our final estimation. The idea can be formalized as:
\begin{equation}
	\label{equ1}
	I_{final} = I_{l}\times W_{l}+ I_{m}\times W_{m} + I_{s}\times W_{s}
\end{equation}
where $I_{l}$, $I_{m}$, $I_{s}$ represent the illuminant distribution maps estimated from large-, medium- and small-scale images respectively, while $W_{l}$, $W_{m}$, $W_{s}$ denote their corresponding pixel-wise weight maps. These weight maps determine the pixel-wise weights for these illuminant distribution maps, enabling adaptive fusion of multi-scale illuminant maps.

\begin{figure*}[htbp]
	\centering
	\includegraphics[width=1\textwidth]{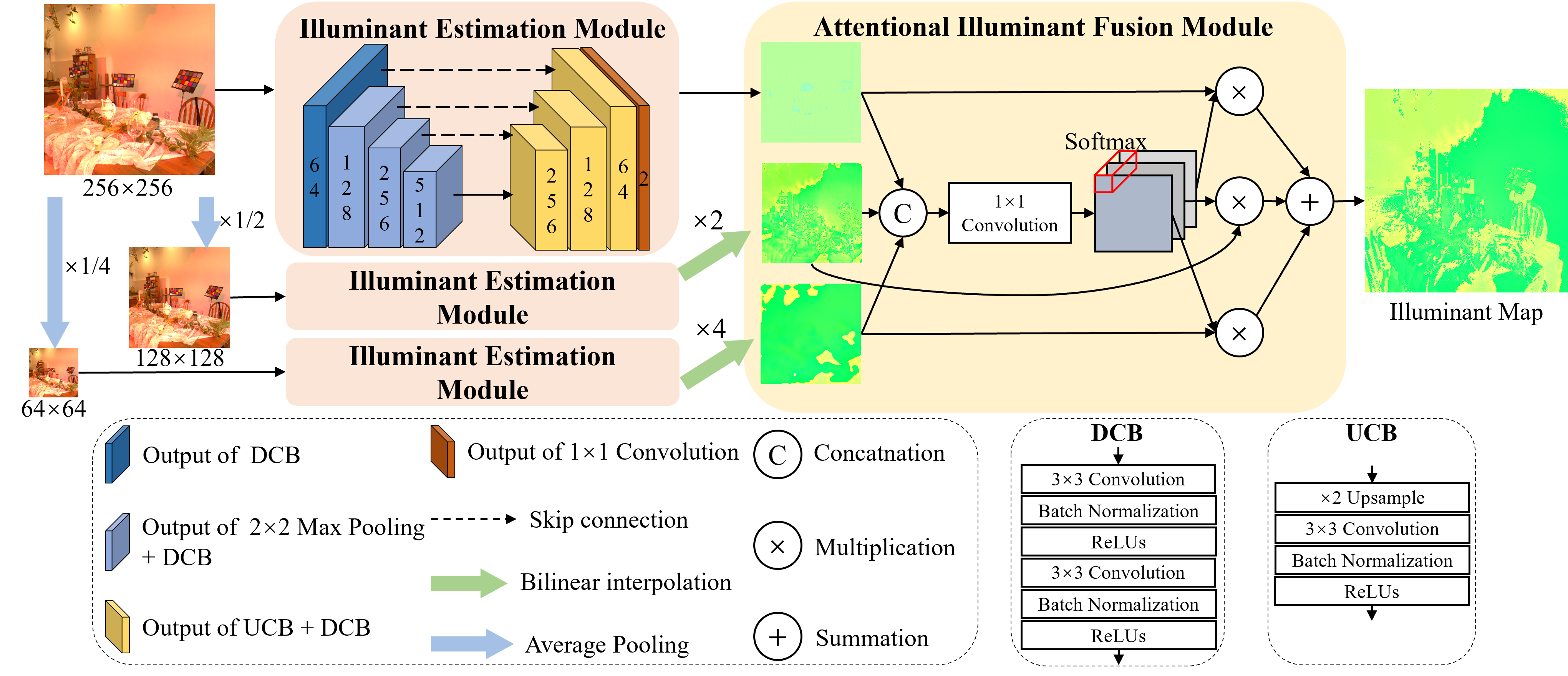} 
	\caption{The structure of our framework.}
	\label{fig1}
\end{figure*}

Following the mathematical formulation, we propose a multi-scale illuminant estimation framework, as shown in Fig. \ref{fig1}, which estimates illuminant maps from images of multiple scales and fuse them linearly. The framework consists of three parallel branches, each equipped with an illuminant estimation module (IEM) to estimate the illuminant distribution from an image of a certain scale. These illuminant distributions are subsequently merged in the attentional illuminant fusion module (AIFM), which can automatically determine the pixel-wise weight maps for these illuminant distributions.

\subsection{Illuminant Estimation Module}
\label{sec3.2: Illuminant Estimation Module}
The structure of IEM is mainly taken from LSMI-U, an variant of U-Net, and its basic components are double convolution blocks (DCB) and upsampling convolution blocks (UCB). A DCB contains two convolution layers, both followed by batch normalization and ReLUs nonlinearity. The encoding pathway consists of four cascaded DCBs, interleaved with three max-pooling layers that progressively halve spatial dimensions. In the decoding pathway, three UCBs are used to double the spatial dimension of features. The output of each UCB is concatenated with features skipped from the encoding pathway and their combination are further processed by a DCB for low- and high-level information fusion. Each illuminant estimation module outputs an illuminant distribution map with only red and blue channels, as the green channel is 1 by default.

\subsection{Attentional Illuminant Fusion Module}
\label{sec3.3: Attentional Illuminant Fusion Module}
The attentional illuminant fusion module takes as input three illuminant distribution maps estimated by three IEMs and adaptively fuses them through an attention mechanism. This module initially concatenates these illuminant distribution maps along the channel dimension, followed by a convolution layer to generate a tensor with three channels. This tensor is subsequently normalized across the channel dimension with a softmax function, producing three weight maps that quantitatively characterize the pixel-wise importance of each illuminant maps. Finally, following Eq. \ref{equ1}, the three weight maps and the corresponding illuminant distributions are linearly combined to produce the final output.

\subsection{Loss Function}
\label{sec3.4: Loss Function}
Mean angular error (in degree) is used to measure the error between the model output $I_{output}$ and the ground truth illuminant map $I_{gt}$:

\begin{equation}
	\label{equ2}
	\frac{1}{N}\sum_{i=1}^{N}\arccos({\frac{I_{output}(i) \cdot I_{gt}(i)}{\|I_{output}(i)\|_{2}\|I_{gt}(i)\|_{2}}})
\end{equation}
where $\|\cdot\|_{2}$ denotes the $L_{2}$ norm, $i$ represents the pixel index and $N$ represents the total number of pixels.

\subsection{Implementation Details}
Our method is implemented with Pytorch \cite{paszke_pytorch_2019} and runs on a platform with Python 3.12, Nvidia RTX4090, CUDA version 12.2 and Ubuntu 22.04 operating system. We use AdmW \cite{loshchilov_decoupled_2019} to optimize our model for 600 epochs with a mini-batch size of 16. The initial learning rate is 0.001 and linearly decays for the first 400 epochs:
\begin{equation}
	\label{equ3}
	\frac{600-epoch}{600}\times0.001, \quad epoch <=400
\end{equation}
During the last 200 epochs, the learning rate remains constant to facilitate convergence. 

\section{Experiments and Discussion}
\label{sec4:experiment}
\subsection{Dataset and Error Metric}
\label{sec4.1}
\begin{figure}[htbp]
	\centering
	\includegraphics[width=0.5\textwidth]{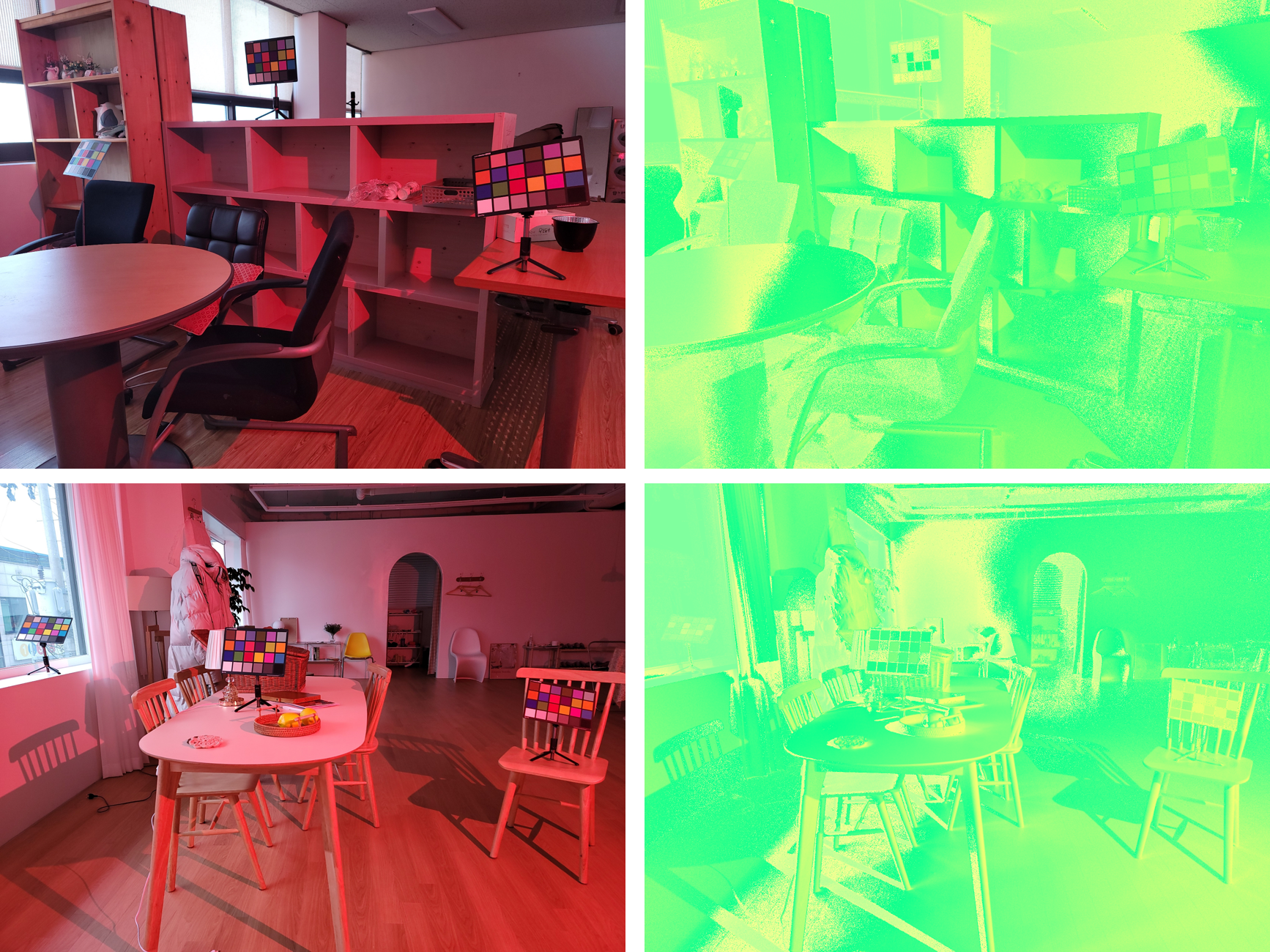} 
	\caption{Sample images from LSMI dataset. Multi-illuminant images (left). Pixel-wise illuminant maps (right).}
	\label{fig2}
\end{figure}

We perform our experiments on the LSMI dataset, specifically designed for developing and evaluating multi-illuminant color constancy methods. The dataset contains three subsets captured respectively by Samsung Galaxy Note20 Ultra, Nikon D810, Sony $\alpha$9. There are totally 7,486 high-quality images. Each image is labeled with pixel-wise mixing ratios of its interior illuminants and illuminant chromaticities, by which we can calculate the ground truth (GT) illuminant distribution maps. Some example images are shown in Fig. \ref{fig2}. It has to be noted that the ground truth illuminant maps are normalized by the green channel, so the greenish color. 

During our experiments, each subset is randomly divided into training, validation and test sets with a ratio of 7:2:1. We uses linear Tiff images as input and resize them to the size of $256\times256$. The mean angular error, as described in Eq. \ref{equ2}, is used as our error metric. The performance of a given method is quantitatively evaluated with four statistical metrics: mean, standard deviation (SD), median, and trimean values of the angular errors across a dataset. Lower values across these metrics indicate superior performance.

\subsection{Parameter Setting}
\label{sec4.2}
First of all, we perform a parameter sensitivity experiment to determine the optimal parameter setting for our method. Based on the baseline model described in Section \ref{sec3: method}, we extend the model depth by symmetrically adding more convolutional layers to the IEM. Additionally, we introduce a multiplier $\theta$ to control the number of channels of output tensors throughout IEM. This experiment is performed on the Galaxy subset. We train several models with distinct parameter settings on the training set and validate their performance on the validation set. Quantitative results for different parameter settings are presented in Table \ref{tab1}, with the best results highlighted in bold. These experimental results demonstrate that our model performs best with 8 convolutional layers in the IEM and $\theta = 1$.

\begin{table}[htb]
	\caption{Results of different parameter settings on the Galaxy subset.}
	\centering
	\resizebox{0.6\columnwidth}{!}{
		\begin{tabular}{cc|ccc}
			\hline
			Layer & $\theta$ & Mean $\downarrow$          & Median $\downarrow$       & Trimean $\downarrow$      \\ \hline
			12     & 1/4         & 1.92          & 1.48          & 1.56          \\
			12     & 1/2          & 1.90          & 1.49          & 1.54         \\
			12     & 1          & 1.87          & 1.45          & 1.54          \\
			8     & 1/4          & 1.94          & 1.49          & 1.60         \\
			8     & 1/2          & 1.83          & 1.43          & 1.50         \\
			8     & 1          & \textbf{1.74} & \textbf{1.28} & \textbf{1.39} \\ \hline
		\end{tabular}
	}
	\label{tab1}
\end{table}

\subsection{Ablation Study}
Our model mainly consists of three IEMs and one AIFM, so we conduct an ablation study to validate the necessity of these modules. We create some variants by removing individual modules from the baseline model. These variants are trained and evaluated respectively on the training and validation set of the Galaxy subset. Table \ref{tab2} presents the results of this experiment and the best results are highlighted in bold. As the table shows, the baseline model performs best compared to these variants, indicating the necessity of all modules. 

\begin{figure}[htbp]
	\centering
	\includegraphics[width=0.8\textwidth]{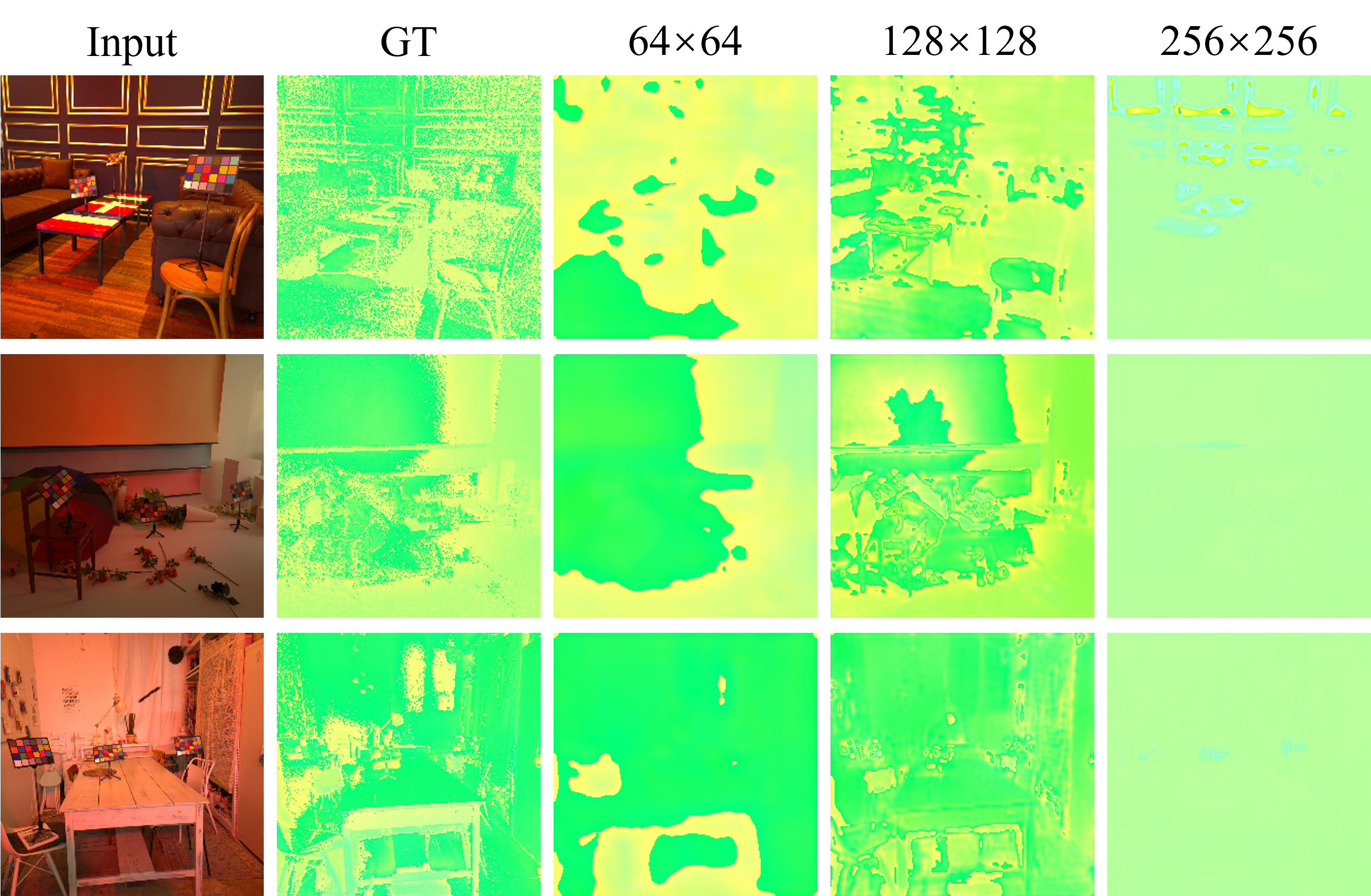} 
	\caption{Illuminant maps estimated from different scales. }
	\label{fig3}
\end{figure}

To further analyze the distinct roles of three IEMs, we visualize the intermediate outputs of our model. Fig. \ref{fig3} demonstrates three sample images, their corresponding ground truth illuminant maps and the illuminant maps estimated from different scales. As shown in Fig. \ref{fig3}, the illuminant maps estimated from small-scale images are smoother, which characterize the coarse-grained illuminant distributions. Illuminant maps estimated from medium scales reveal finer structural details, while large-scale estimations contain only fine-grained details. These results demonstrate that our three-branch model can capture complementary features from multi-scale images, enabling our model to restore the illuminant map accurately.

\begin{table}[htb]
	\caption{Results of ablation study on the Galaxy subset.}
	\centering
	\resizebox{0.7\columnwidth}{!}{
		\begin{tabular}{cccc|ccc}
			\hline
			256 & 128 & 64 & AIFM & Mean $\downarrow$          & Median $\downarrow$        & Trimean $\downarrow$       \\ \hline
			-   & \checkmark   & \checkmark  & \checkmark   & 1.79          & 1.33          & 1.41          \\
			\checkmark   & -   & \checkmark  & \checkmark   & 1.79          & 1.32          & 1.43          \\
			\checkmark   & \checkmark   & -  & \checkmark   & 1.85          & 1.41          & 1.51          \\
			\checkmark   & \checkmark   & \checkmark  & -   & 1.75          & 1.36          & 1.44          \\
			\checkmark   & \checkmark   & \checkmark  & \checkmark   & \textbf{1.74} & \textbf{1.28} & \textbf{1.39} \\ \hline
		\end{tabular}
	}
	\label{tab2}
\end{table}

\subsection{Performance Comparison}
We have compared our method with state-of-the-art methods both quantitatively and qualitatively. All methods are trained, if applicable, and tested on the three subsets of LSMI. The color checkers of these test sets are masked out to avoid potential data bias. Quantitative results are shown on Table \ref{tab3}, with the best results highlighted in bold. As the table shows, our method performs best among these methods and significantly surpasses existing methods. For example, on the Galaxy subset, the mean error of our method is $1.96^{\circ}$ which is $12\%$ lower than the second best result $2.23^{\circ}$.

\begin{figure*}[htb]
	\centering
	\includegraphics[width=1\textwidth]{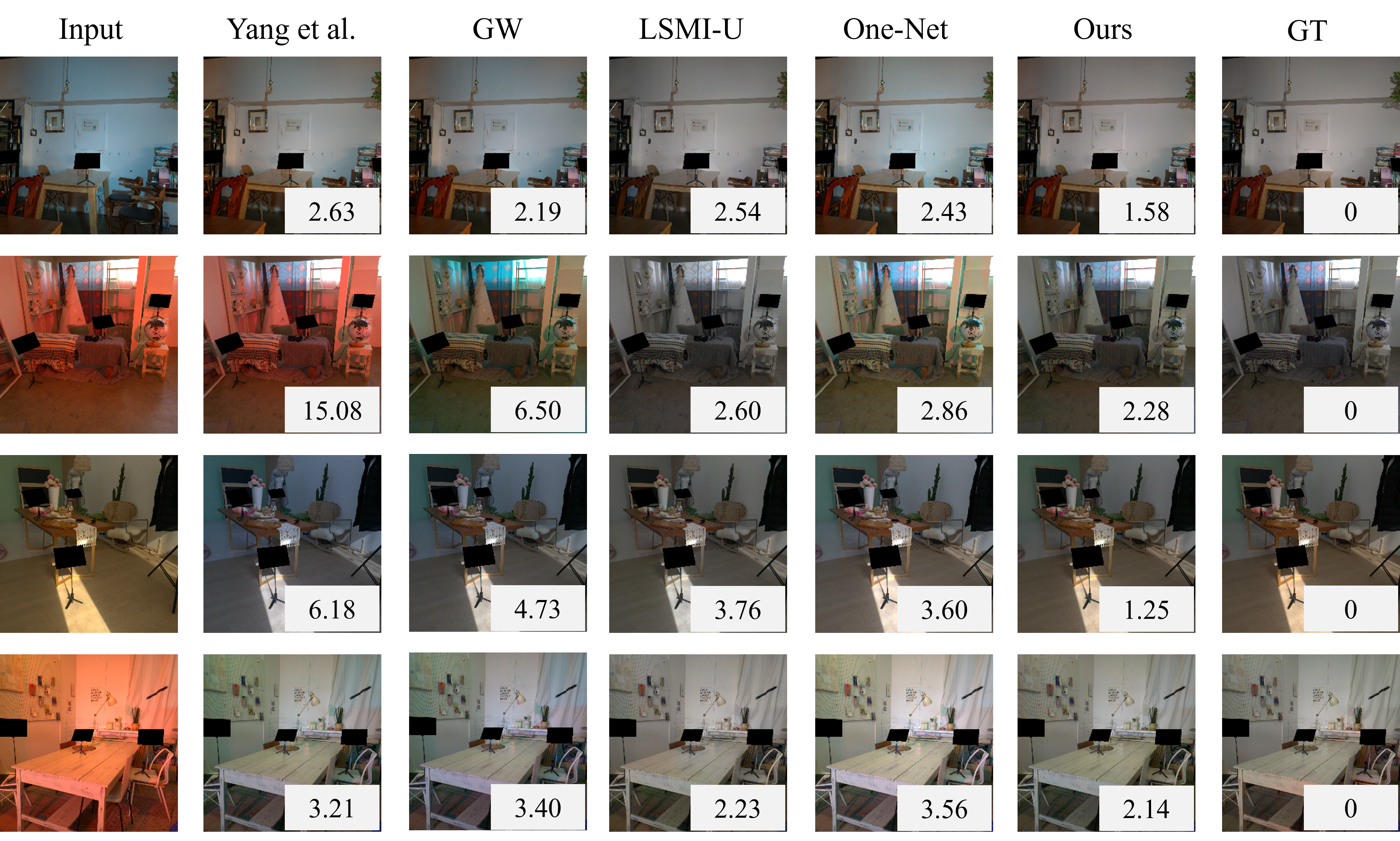}
	\caption{Qualitative comparison between different methods.}
	\label{fig4}
\end{figure*}

Fig. \ref{fig4} shows some images corrected by different methods for qualitative comparison. The mean angular errors of these images are presented on the bottom right corner. As we can see, the local color biases are corrected by our method and our results are visually closer to the ground truth. 

\begin{table*}[h]
	\caption{Quantitative comparison on LSMI. "*" indicates that the method requires training.}
	\centering
	\resizebox{\textwidth}{!}{
		\begin{tabular}{c|cccc|cccc|cccc}
			\hline
			\multirow{2}{*}{Name} & \multicolumn{4}{c|}{Galaxy}                                   & \multicolumn{4}{c|}{Nikon}                                    & \multicolumn{4}{c}{Sony}                                      \\ \cline{2-13} 
			& Mean$\downarrow$          & SD$\downarrow$           & Median$\downarrow$        & Trimean$\downarrow$       & Mean$\downarrow$          & SD$\downarrow$           & Median$\downarrow$        & Trimean$\downarrow$       & Mean$\downarrow$          & SD$\downarrow$           & Median$\downarrow$        & Trimean$\downarrow$       \\ \hline
			GW \cite{Buchsbaum1980a}           & 4.67          & 3.21          & 4.31          & 4.32          & 3.43          & 2.23          & 2.96          & 3.02          & 4.53          & 2.59          & 4.12          & 4.20          \\
			WP \cite{barnard_comparison_2002}               & 12.12         & 6.33          & 15.01         & 13.38         & 9.72          & 6.34          & 9.75          & 9.57          & 10.46         & 7.41          & 9.15          & 9.99          \\
			SoG \cite{finlayson_shades_2004}         & 8.08          & 5.63          & 7.38          & 7.75          & 6.80          & 5.03          & 5.99          & 6.30          & 6.89          & 5.19          & 5.52          & 5.89          \\
			GE \cite{VandeWeijer2007c} ($n=1, p=7, \sigma=4$)        & 8.03          & 6.34          & 6.06          & 7.18          & 6.90          & 5.50          & 5.36          & 6.11          & 6.67          & 5.77          & 4.46          & 5.21          \\
			GE \cite{VandeWeijer2007c} ($n=2, p=7, \sigma=4$)           & 7.99          & 6.09          & 5.98          & 7.08          & 7.14          & 5.53          & 5.50          & 6.28          & 6.85          & 6.00          & 4.61          & 5.50          \\
			Cheng et al. \cite{Cheng2014a}             & 7.25          & 6.13          & 4.71          & 6.09          & 5.53          & 4.56          & 4.16          & 4.84          & 5.59          & 4.99          & 3.88          & 4.26          \\
			Yang et al. \cite{Yang2015}      & 3.99          & 4.06          & 2.77          & 2.95          & 3.78          & 3.92          & 2.69          & 2.83          & 4.14          & 3.15          & 3.26          & 3.53          \\
			Qian  et al. \cite{qian_finding_2019}      & 5.05          & 5.60          & 2.84          & 3.33          & 3.53          & 3.43          & 2.48          & 2.74          & 4.77          & 4.92          & 3.26          & 3.55          \\
			Weighted GE \cite{gijsenij_improving_2012}            & 8.64          & 6.75          & 6.20          & 7.42          & 6.60          & 6.29          & 4.53          & 5.04          & 7.94          & 7.19          & 5.08          & 6.07          \\
			LSMI-U* \cite{kim_large_2021}                & 3.09          & 2.28          & 2.34          & 2.47          & 2.36          & \textbf{1.36} & 1.98          & 2.07          & 3.53          & 1.59          & 3.30          & 3.33          \\
			One-Net* \cite{domislovic_color_2023}               & 2.23          & 2.04          & 1.74          & 1.84          & 2.34          & 1.73          & 1.90          & 2.02          & 2.33          & 1.55          & 2.02          & 2.08          \\
			Ours*                   & \textbf{1.96} & \textbf{1.66} & \textbf{1.50} & \textbf{1.60} & \textbf{1.90} & 1.46          & \textbf{1.44} & \textbf{1.55} & \textbf{1.76} & \textbf{1.27} & \textbf{1.51} & \textbf{1.56} \\ \hline
		\end{tabular}
	}
	\label{tab3}
\end{table*}

\section{Conclusion}
\label{sec5:Conclusion}
In this paper, we propose that an illuminant distribution map can be represented as the linear combination of multi-grained illuminant distribution maps. Based on this point of view, we propose a multi-scale illumination distribution estimation and fusion framework. Considering that larger-scale images contain more detailed information, we construct a three-branch convolutional networks, each branch implemented using a U-Net. These branches estimate multi-grained illumination distribution maps from large-, medium- and small-scale images. An attentional illumination fusion module adaptively generates pixel-wise weight maps to linearly combine these multi-granularity distributions. Through parameter sensitivity experiment, we determine the optimal parameter setting for our method. Ablation studies are conducted to validate the necessity of each module and analyze the distinct roles of three branches. Comparative experiments demonstrate that our approach achieves state-of-the-art performance.

\section{Acknowledgement}
\label{sec:ack}
This work was supported by Engineering Research Center of Hubei Province for Clothing Information, Wuhan Textile University (2024HBCI03).

\bibliography{refs} 
\bibliographystyle{spiebib} 

\end{document}